# Class-independent sequential full image segmentation, using a convolutional net that finds a segment within an attention region, given a pointer pixel within this segment

Sagi Eppel[1]

## Abstract

This work examines the use of a fully convolutional net (FCN) to find an image segment, given a pixel within this segment region. The net receives an image, a point in the image and a region of interest (RoI ) mask. The net output is a binary mask of the segment in which the point is located. The region where the segment can be found is contained within the input RoI mask. Full image segmentation can be achieved by running this net sequentially, region-by-region on the image, and stitching the output segments into a single segmentation map. This simple method addresses two major challenges of image segmentation: 1) Segmentation of unknown categories that were not included in the training set. 2) Segmentation of both individual object instances (things) and non-objects (stuff), such as sky and vegetation. Hence, if the pointer pixel is located within a person in a group, the net will output a mask that covers that individual person; if the pointer point is located within the sky region, the net returns the region of the sky in the image. This is true even if no example for sky or person appeared in the training set. The net was tested and trained on the COCO panoptic dataset and achieved 67% IOU for segmentation of familiar classes (that were part of the net training set) and 53% IOU for segmentation of unfamiliar classes (that were not included in the training). Code has been made available at this [URL](URL).

## 1. Introduction

The segmentation of images into regions corresponding to different things is one of the main challenges in computer vision[1,2]. Splitting an image to different segments can be done at several different levels. The two most common segmentation approaches are: 1) Semantic segmentation, in which the image is split into different regions corresponding to different semantic categories (sky, grass, cars etc)[1,2]; and 2) instance segmentation, which involves splitting the image into regions corresponding to individual objects (such as a single person in a crowd)[2,3].

One of the main challenges in image segmentation is to find a single segmentation method that is able to separate an arbitrary image into regions corresponding to object instances ('things')[4], and that is also able to segment non-object regions ('stuff') such as the ocean or sky[4-6,] and to achieve this for images containing unfamiliar classes that were not encountered during the training stage. This work demonstrates how the use of a fully convolutional net (FCN)[2] to solve a much simpler segmentation problem can also solve the general segmentation problem described above. Firstly, we consider the simple segmentation task of finding a region corresponding to a single segment within an image, given a point within this segment (Figure 1). An additional optional input is a region of interest (RoI) mask that contains the region of the image within which the output segment may be (Figure 1)[7].

---

[1] sagieppel@gmail.com Vaya Vision

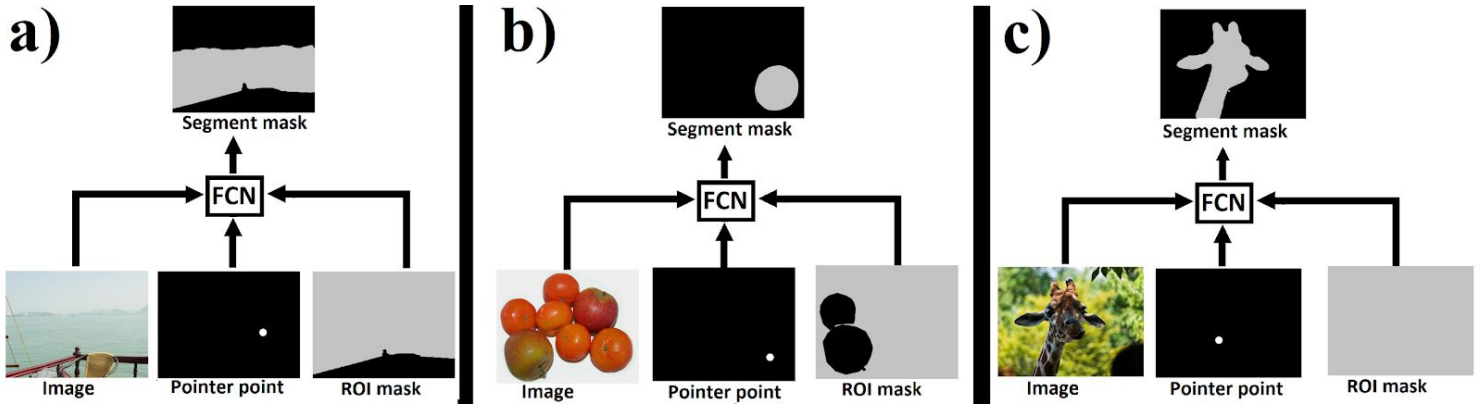

**Figure 1: A fully convolutional neural net (FCN) receives an image, a pointer pixel within the image and an RoI mask, and returns a mask of the segment in which the pointer is located (The segment is contained within the RoI mask). The figures show real results for the segmentation of unfamiliar classes that were not included in the training set. All images were taken from the COCO dataset[4].**

For example, if the pointer pixel is located within a fruit in a group (Figure 1b), the output segment will be the region of that individual fruit (Figure 1b), while if the pointer point is located in the ocean, the net will return the ocean region in the image (Figure 1a). This should be the case even if fruits and oceans did not appear in any of the training images. Although this method is restricted to finding a single segment at a time, it can easily be used for full image segmentation by applying it sequentially and removing the output segment predicted at each step from the RoI mask used as input for the next step (Figure 2).

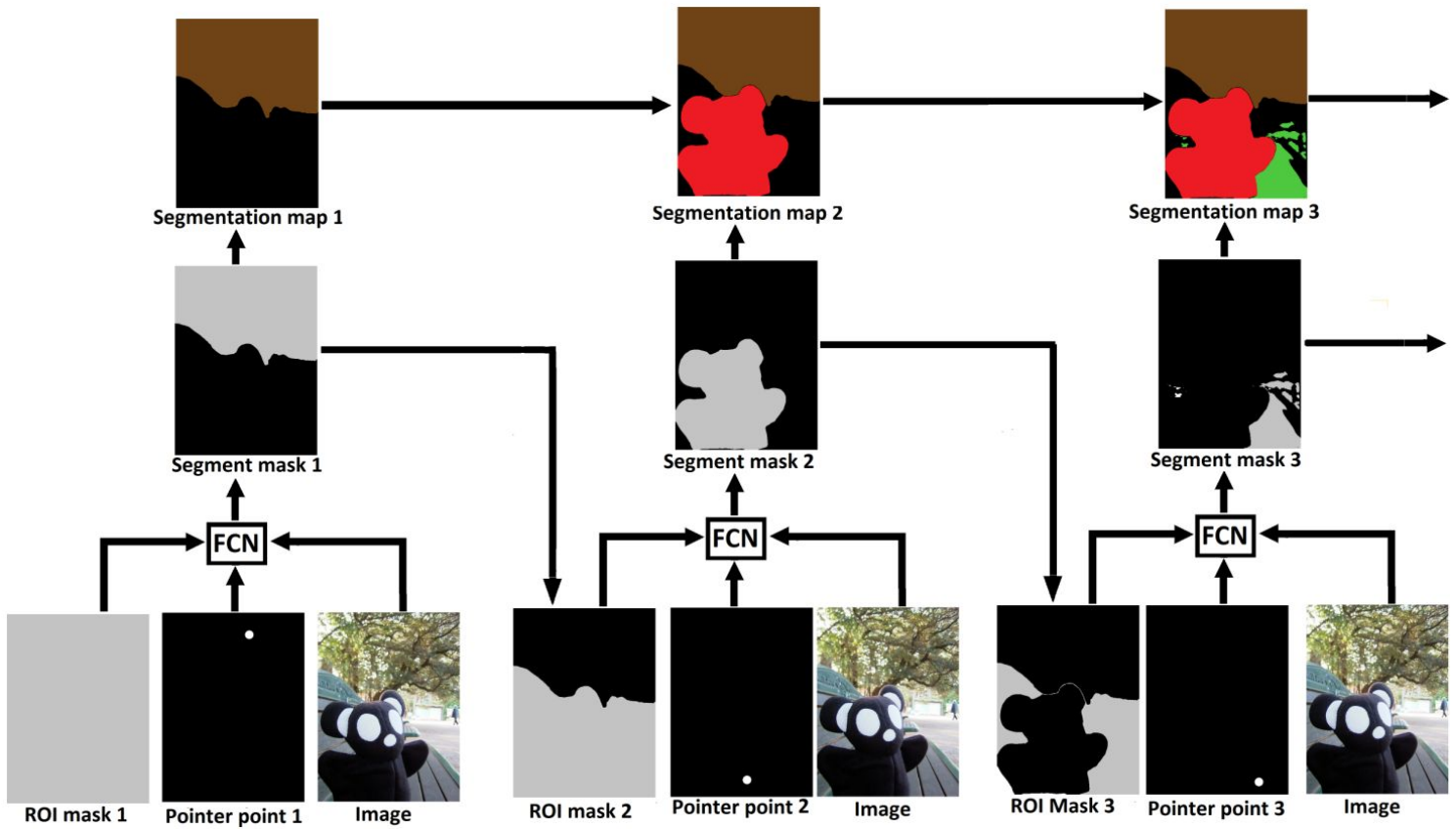

**Figure 2.a: Region-by-region full image segmentation using pointer segmentation approach.**

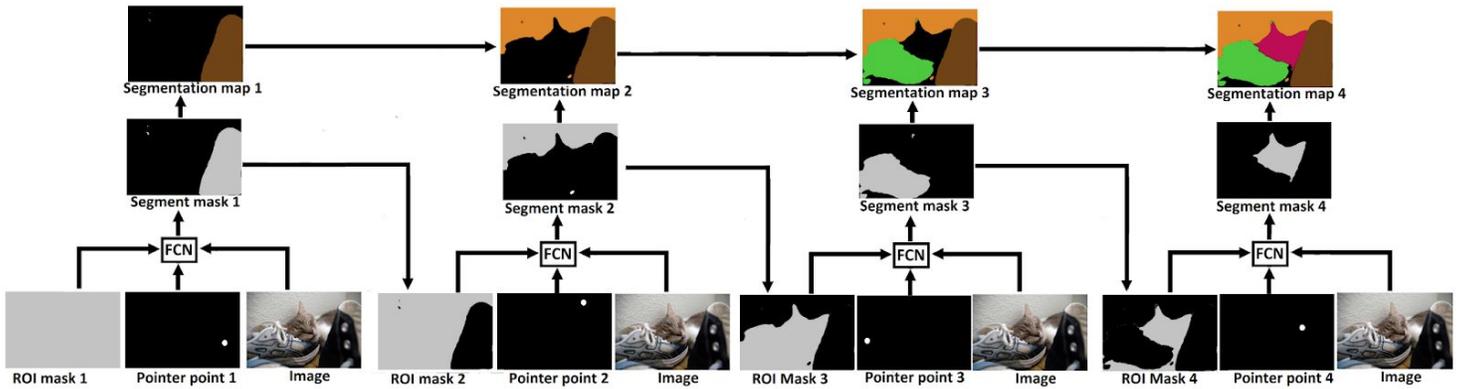

**Figure 2b:** Sequential full image segmentation, using a pointer segmentation: i) an RoI mask that covers the entire image is generated; ii) a random pointer pixel is selected within the new RoI mask; iii) the image RoI mask and the pointer pixel are used as input to the FCN; iv) the predicted segment mask is added to the segmentation map and removed from the RoI mask. Steps 2-4 are repeated with the new RoI mask.

The method is trained on a large number of segments corresponding to different categories, and returns the region of the segment as a binary mask, regardless of the segment class. As result, the method learns a general class-agnostic segmentation pattern and can segment even unknown categories that were not encountered during the training stage. In addition, since the method is class-independent, it can easily achieve segmentation of objects (things), such as individual animals or fruits (Figure 1b,c), as well as non-object classes (stuff) such as the sky and ground (Figure 1a).

## 2. Related works

State-of-the-art methods for image segmentation currently revolve around machine learning with deep neural nets[1]. Machine-learning-based approaches for image segmentation are mostly based on the use of fully convolutional networks (FCNs) to recognize a specific set of pre-given classes[1-3]. This FCN can then segment the image into regions of different semantic classes by assigning a class to each pixel in the image. The main limitation of this semantic segmentation approach is that it cannot separate two object instances of the same class, such as two people standing side by side. To resolve this, several methods for instance-aware segmentation have been suggested[1,3]. Mask R-CNN is one method that performs instance segmentation by assigning a bounding box for each object and then finding the mask of the object instance within each bounding box[3]. While this approach can be trained to segment object instances even with unknown classes[8], it is limited for the segmentation of objects, and cannot segment non-object classes. An alternative sequential approach for segmenting object instances, one segment at a time, is based on recurrent neural networks (RNN)[9-11]. This method is again limited to finding only instances of specific classes of objects. Another approach that is closely related to the method examined here is interactive segmentation[12-15]. In this case, a person guides the segmentation process by marking several points or scribbles within the image region corresponding to the object to be segmented. These markings are then passed as an additional input to the FCN, which returns the marked object segment as a binary mask. Another semi-sequential method that has some similarity to the method described here is based on a sliding window approach; this scans the image in patches and finds the mask of the segment that corresponds to each patch[16-18].

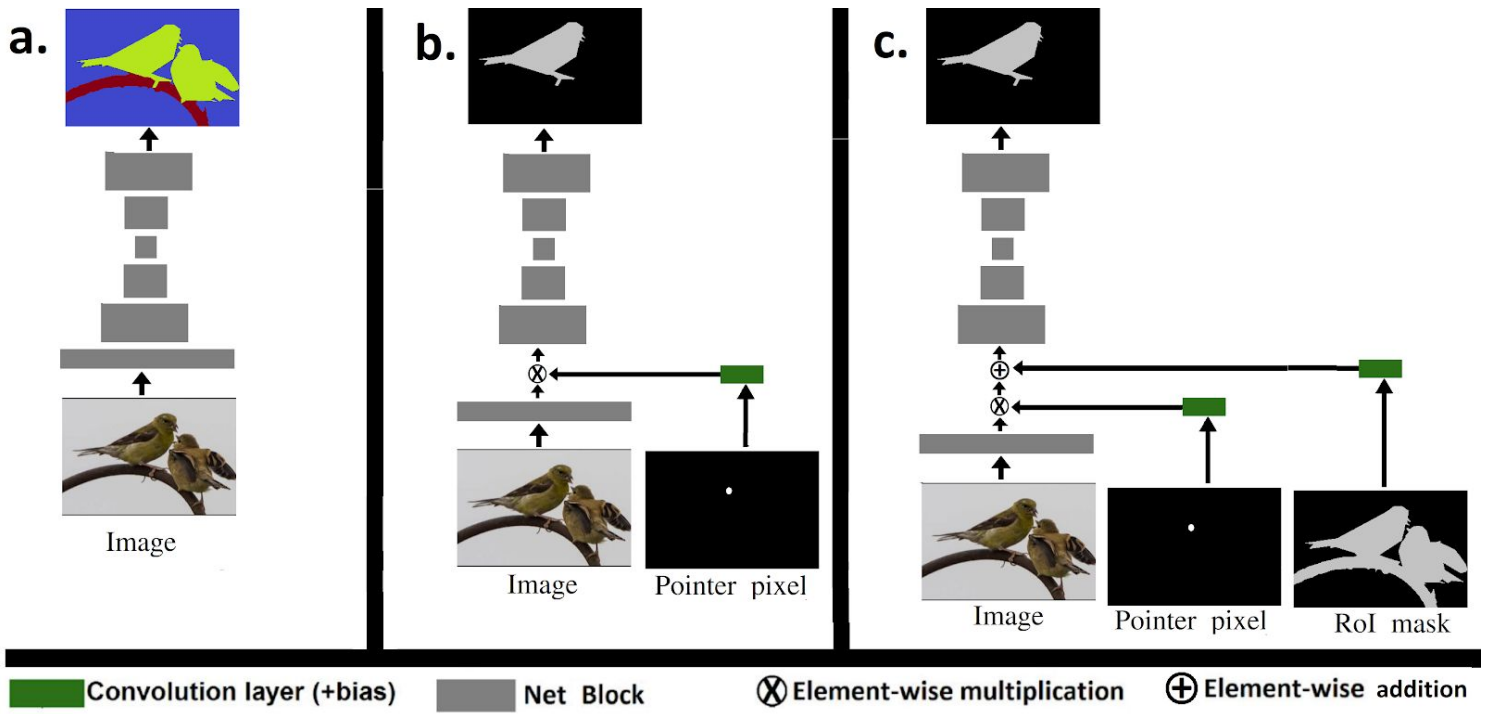

Figure 3: a) Standart FCN for semantic segmentation b) FCN for finding a single segment given a pointer pixel. c) FCN for finding a single segment given a pointer and an RoI mask.

## 3. Net structure

Finding the image segment that contains a given pointer point (Figure 1) can be easily achieved using a fully convolutional neural network (FCN). While a standard FCN receives only an image as input (Figure 3.a), it is possible to introduce additional inputs by converting them to matrix/image format. The pointer pixel can be converted to image form by generating a single-channel image in which the pointer pixel is set to one and the value of all other pixels is set to zero (Figure 3.b). This pointer image is then processed using a single convolutional layer to produce an attention map, which has the same shape as the activation map generated from the RGB image after the first convolutional layer (Figure 3.b). The attention map and the image activation map are merged by element-wise multiplication. The resulting merged map is fed to the next layer of the FCN, which remains the same as in a standard FCN. The merging of the attention map in the first layers of the net has been shown to be superior to merging in higher layers[7]. The addition of an RoI mask as further input can also be achieved by representing this as a binary map in which the values of pixels within the RoI are set to one, and the rest to zero (Figure 3.c). This map is again used as an input to the net by passing it through a single convolutional layer and merging the resulting attention map with the previous merged layer using element-wise addition (Figure 3.c). The net used for this was a standard FCN with a Resnet encoder, followed by PSP layers and three upsampling layers[19,20]. Code has been made available at this URL.

## 4. Training and evaluation

The net was trained on the COCO panoptic dataset[4]. This dataset contains images that are annotated with both instance-level segmentation (things) and non-object classes (stuff). Altogether, the dataset contains 180 different semantic classes. Training the net on a large number of different classes is essential in order to achieve general class-independent segmentation properties. Training was carried out by randomly selecting a single segment from the annotation map of a given image. A single pixel was randomly chosen within this segment and used as the input pointer point. To generate an RoI mask for the training, a random number of additional segments was chosen from the annotation map and their regions were excluded from the RoI mask (by setting any region contained within these segments to zero in the RoI mask). The net was trained with this type of RoI mask for 80% of the iterations, while the remaining 20% were carried out with an RoI mask that covered the entire image. The segments for training were randomly selected with a probability proportional to their areas. If the selected segment corresponded to an object class (thing), the entire region of the individual object instance was chosen. For non-object (stuff) classes, the segment was generated by taking a connected region of the pixels corresponding to the same class, using the connected component method. In order to test the net on unfamiliar classes, 20% of the categories of the COCO dataset were excluded from the training and were used only in the evaluation stage. Evaluation of the accuracy of this net was carried out using the validation subset of the COCO panoptic dataset. The segment and input selection methods were identical to the one used in the net training. The statistics were based on intersection over union (IOU).

## 5. Segmentation of familiar and unfamiliar classes

To evaluate the accuracy of the net on unfamiliar classes, the net was trained on 80% of the classes in the COCO panoptic training set and evaluated on the remaining 20% classes in the COCO validation set. The results are given in Table 1 and Figure 4 (for unfamiliar classes). From Table 1.a it can be seen that the average IOU for segmentation of unfamiliar classes is 53% (for cases with an RoI that covers the entire image), while that for a familiar class is 67%. Hence, the net achieves reasonable accuracy in the segmentation of unfamiliar classes that were not encountered during training, although it still achieves better accuracy for familiar classes on which it was trained. As shown in Figure 4 the main type of errors in the segmentation of unfamiliar classes is tracing a single part of the object as the complete object or marking two connected objects as a single object. A comparison of the accuracy of 'things' vs. 'stuff' (objects vs. non-objects) shows that despite the differences between these two categories, they gave for most cases similar segmentation accuracies in terms of the average IOU (Table 1). Hence, it is clear that the single net has learned to segment both object instances and non-object classes, and that it can do this even for unfamiliar classes that were not included at the training stage. As can be expected the accuracy of the segmentation is better for large segments (Table 1.b) and for RoI mask that covers a small fraction of the image area (Table 1.a).

Table 1. Segmentation Intersection Over Union (IOU) for single segment prediction (Figure 1) using COCO panoptic evaluation dataset.

a) IOU with respect to RoI mask as a fraction of the image.   b) IOU with respect to the segment size in pixels.

| a | Unfamiliar Classes[1] | | | Familiar classes[2] | | | b | Unfamiliar Classes[1] | | | Familiar classes[2] | | |
|---|---|---|---|---|---|---|---|---|---|---|---|---|---|
| RoI as Image Fraction< | IOU all | IOU things[3] | IOU stuff[4] | IOU all | IOU things[3] | IOU stuff[4] | Segment Size (pixels) | IOU all | IOU things[3] | IOU stuff[4] | IOU all | IOU things[3] | IOU stuff[4] |
| 10% | 73% | 72% | 73% | 79% | 80% | 77% | 1,000 | 24% | 25% | 23% | 26% | 26% | 24% |
| 20% | 72% | 72% | 72% | 80% | 80% | 78% | 2,000 | 35% | 37% | 31% | 41% | 43% | 33% |
| 30% | 69% | 69% | 69% | 77% | 79% | 75% | 4,000 | 45% | 47% | 42% | 51% | 55% | 39% |
| 40% | 66% | 66% | 66% | 76% | 77% | 75% | 8,000 | 50% | 52% | 48% | 59% | 62% | 48% |
| 50% | 63% | 62% | 64% | 74% | 75% | 73% | 16,000 | 57% | 59% | 55% | 65% | 68% | 57% |
| 60% | 62% | 58% | 64% | 73% | 74% | 71% | 32,000 | 63% | 64% | 62% | 72% | 75% | 64% |
| 70% | 60% | 56% | 62% | 71% | 73% | 68% | 64,000 | 68% | 69% | 68% | 77% | 80% | 72% |
| 80% | 58% | 53% | 61% | 69% | 69% | 68% | 128,000 | 69% | 67% | 71% | 78% | 80% | 75% |
| 90% | 56% | 51% | 58% | 69% | 68% | 70% | 256,000 | 62% | 60% | 63% | 76% | 77% | 75% |
| 100% | 53% | 49% | 56% | 67% | 67% | 67% | 500,000 | 59% | 61% | 58% | 81% | 77% | 83% |

1. Unfamiliar classes did not appear in the training set.  2. Familiar classes that occur in the training set.
3. Things are countable objects instance (cars, dogs).  4. Stuff are non-object classes (sky, floor).

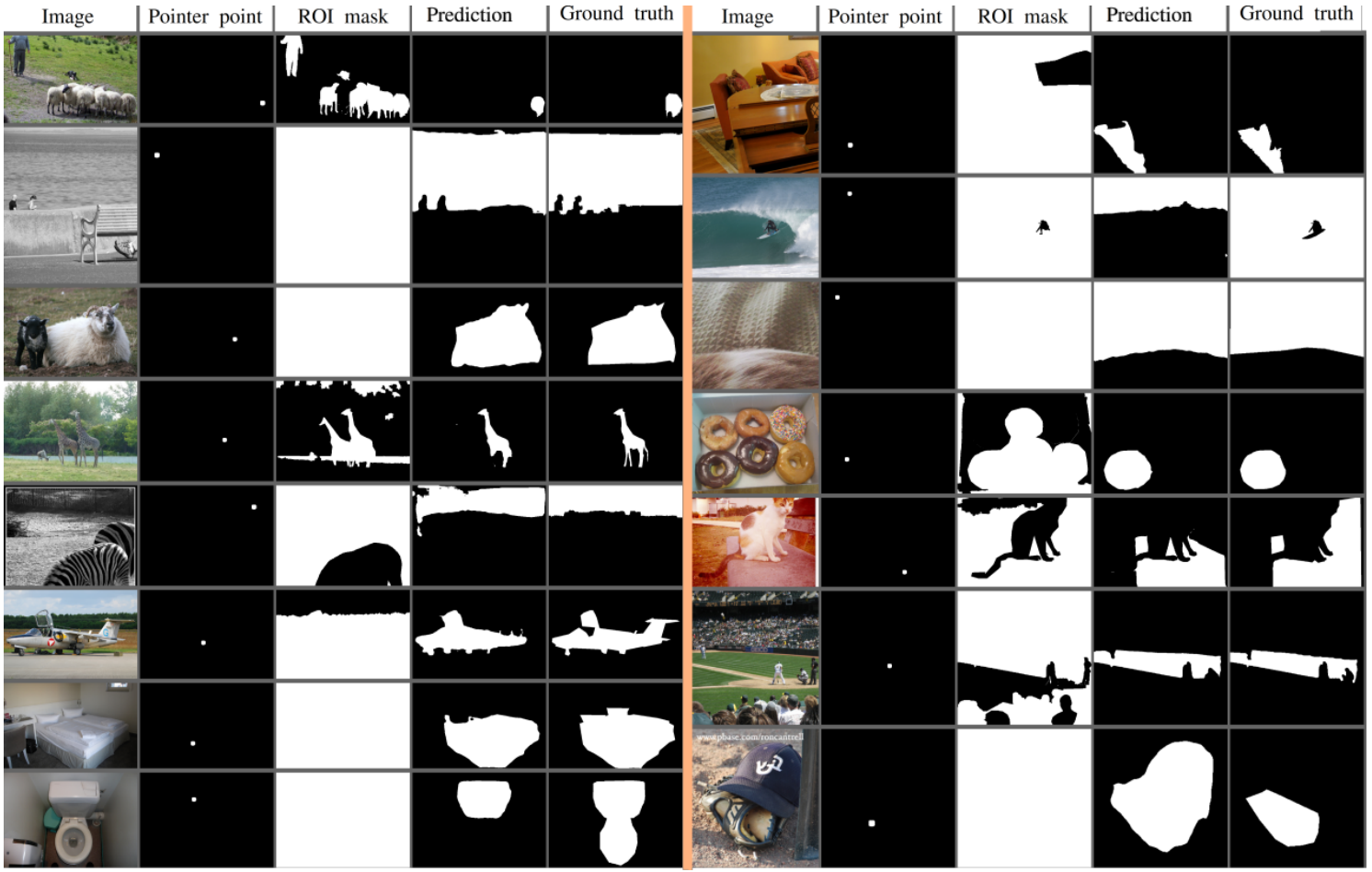

**Figure 4) Result of single segment prediction (Figure 1) for unfamiliar classes that the net never encountered during training.**

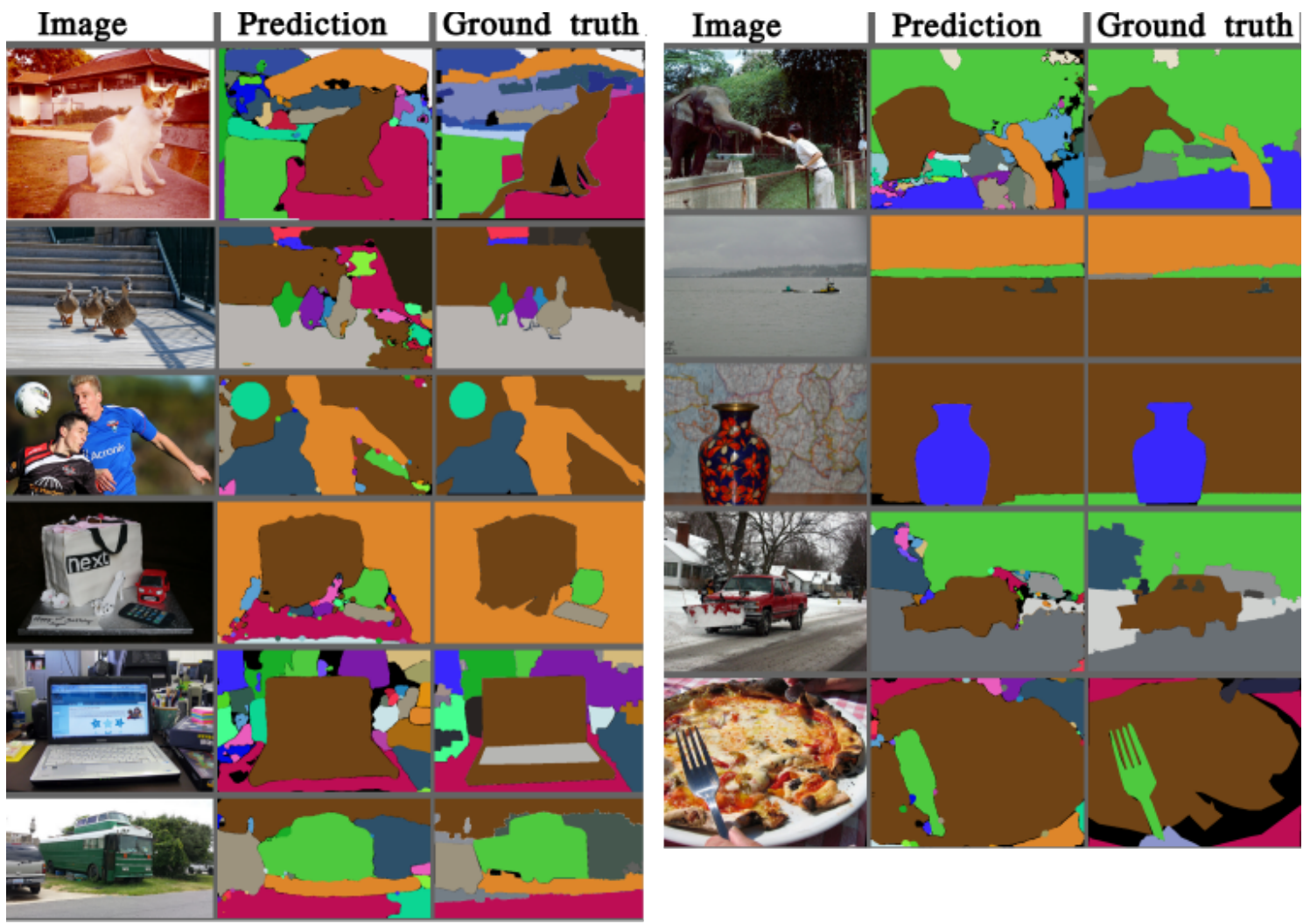

**Figure 5:** Results of full image segmentation using cascade pointer approach (Figure 2). For net that was trained on 80% of the class in the test images.

**Table 2:** Average IOU for full image segmentation using a cascade approach (Figure 2) with RoI and without RoI mask as input to the net. For net that was trained on 80% of the classes in the test images.

|  | IOU all | IOU stuff[2] | IOU things[1] | Precision All | Precision stuff | Precision things | Recall All | Recall stuff | Recall things |
|---|---|---|---|---|---|---|---|---|---|
| With ROI mask | 61% | 60% | 61% | 83% | 85% | 81% | 72% | 69% | 74% |
| No ROI mask | 59% | 58% | 60% | 80% | 83% | 78% | 72% | 69% | 74% |

**1.** Things are countable objects instance (cars, dogs). **2.** Stuff are non-object classes (sky, floor)

# 6. Full image segmentation using a sequential region-by-region approach

The ability of the net to segment full images were examined by running it sequentially, one segment at a time, and then stitching the resulting segments into a single segmentation map. This was done as shown in Figure 2. First, an RoI mask that covered the entire image was generated. A random point in this RoI mask was selected; this point, the image, and the RoI mask were then used as input to the net, and the segment predicted by the net was stitched into the segmentation map. The predicted segment was also removed from the RoI mask (Figure 2). A random point was again selected from the new RoI mask and used along with the new RoI and the image as input to the net. This process was repeated until the segmentation map covered over 95% of the image area. The results of this method are given in table 2 and Figure 5. The accuracy of this method was evaluated by calculating the IOU for each of the segments in the predicted segmentation map with the best-matched segment in the ground truth segmentation map. The results show that the net achieved reasonable accuracy, with an average IOU per pixel of 61% for the entire image (familiar and unfamiliar classes). The same test was run but without the addition of the RoI map as input for the net, and in this case, the resulting average IOU was 59%. Hence, the use of an RoI mask gives a small improvement. It can be seen from Figure 5 that the main types of errors are in small segments and in segments containing distinct parts such as the keyboard of a computer.

# 7. Reference

1. Garcia-Garcia, A., et al. "A review on deep learning techniques applied to semantic segmentation. arXiv 2017." *arXiv preprint arXiv:1704.06857*.
2. Long, Jonathan, Evan Shelhamer, and Trevor Darrell. "Fully Convolutional Networks for Semantic Segmentation." *arXiv preprint arXiv:1411.4038* (2014).
3. He, Kaiming, et al. "Mask R-CNN." *arXiv preprint arXiv:1703.06870* (2017).
4. Kirillov, Alexander, et al. "Panoptic segmentation." *arXiv preprint arXiv:1801.00868* (2018).
5. Li, Yanwei, et al. "Attention-guided unified network for panoptic segmentation." *arXiv preprint arXiv:1812.03904* (2018).
6. de Geus, Daan, Panagiotis Meletis, and Gijs Dubbelman. "Panoptic segmentation with a joint semantic and instance segmentation network." *arXiv preprint arXiv:1809.02110*(2018).
7. Eppel, Sagi. "Setting an attention region for convolutional neural networks using region selective features, for recognition of materials within glass vessels." *arXiv preprint arXiv:1708.08711* (2017).
8. Hu, Ronghang, et al. "Learning to Segment Every Thing." *arXiv preprint arXiv:1711.10370* (2017)
9. Romera-Paredes, Bernardino, and Philip HS Torr. "Recurrent Instance Segmentation." *arXiv preprint arXiv:1511.08250*(2015).
10. Salvador, Amaia, et al. "Recurrent neural networks for semantic instance segmentation." *arXiv preprint arXiv:1712.00617* (2017).
11. Ren, Mengye, and Richard S. Zemel. "End-to-End Instance Segmentation with Recurrent Attention." *arXiv preprint arXiv:1605.09410* (2016).
12. Xu, Ning, et al. "Deep grabcut for object selection." *arXiv preprint arXiv:1707.00243* (2017).
13. Xu, Ning, et al. "Deep Interactive Object Selection." *arXiv preprint arXiv:1603.04042* (2016).

14. Mahadevan, Sabarinath, Paul Voigtlaender, and Bastian Leibe. "Iteratively trained interactive segmentation." *arXiv preprint arXiv:1805.04398* (2018).
15. Hu, Yang, et al. "A Fully Convolutional Two-Stream Fusion Network for Interactive Image Segmentation." *arXiv preprint arXiv:1807.02480* (2018).
16. Pinheiro, Pedro O., Ronan Collobert, and Piotr Dollar. "Learning to Segment Object Candidates." *arXiv preprint arXiv:1506.06204* (2015).
17. Dai, Jifeng, et al. "Instance-sensitive Fully Convolutional Networks." *arXiv preprint arXiv:1603.08678* (2016).
18. Li, Yi, et al. "Fully Convolutional Instance-aware Semantic Segmentation." *arXiv preprint arXiv:1611.07709* (2016).
19. He, Kaiming, et al. "Deep Residual Learning for Image Recognition." *arXiv preprint arXiv:1512.03385* (2015).
20. Lin, Tsung-Yi, et al. "Feature Pyramid Networks for Object Detection." *arXiv preprint arXiv:1612.03144* (2016).